\documentclass[11pt, logo, onecolumn, copyright, colorlinks=true, allcolors=blue]{nvidiatechreport}
\usepackage[round]{natbib}
\usepackage{url}
\usepackage{multirow}
\usepackage{wrapfig}
\usepackage{needspace}

\title{Where the Model Changes Its Mind: Hindsight-Divergence Localization
for Efficient Reinforcement Learning with Verifiable Rewards}

\author[2,*,\textdagger]{Fanchao Chen}
\author[3,*]{Hengyu Fu}
\author[2,4]{Shivaram Venkataraman}
\author[1,3]{Jiantao Jiao}
\affil[1]{NVIDIA}
\affil[2]{University of Wisconsin--Madison}
\affil[3]{University of California, Berkeley}
\affil[4]{ETH Zurich}

\renewcommand{\copyrightext}{\footerfont $^*$Equal contribution. $^\dagger$Corresponding author: \href{mailto:fchen239@wisc.edu}{\texttt{fchen239@wisc.edu}}.\\
\textcopyright\, \the\year{} NVIDIA. All rights reserved.}
\date{}

\begin{document}

\begin{abstract}
\large \textbf{Abstract.}
Group-relative methods for reinforcement learning with verifiable rewards
(RLVR) learn from differences in rollout outcomes. Independently sampling
complete trajectories is costly and does not explicitly explore the decision
space at critical positions. Feedback on a completed
trajectory can reveal which earlier choices the policy reconsiders,
suggesting where to sample alternative continuations. We introduce \textbf{Hindsight-Divergence
Localization (HDL)}, which uses hindsight-induced changes in token
log-likelihoods to select branch points. HDL generates a small number of
complete root trajectories and fills each training group with continuations
from the selected positions under the original task context.
Each continuation reuses its root prefix and contributes policy updates
only through its newly generated suffix, reducing generation cost while
focusing additional exploration and learning on decisions after branching.
Experiments with three models across math, code, and agent tasks show gains
in both rollout efficiency and task performance. Compared with GRPO at
matched group sizes and training steps, HDL yields up to a 2.5$\times$ reduction
in generated tokens and a 1.8$\times$ speedup in rollout wall-clock time.
Despite this reduced generation budget, HDL improves performance across all
three domains, with gains of up to 12.5 points on agent tasks.
\end{abstract}

\maketitle

\section{Introduction}
\label{sec:intro}

Reinforcement learning with verifiable rewards (RLVR) has been widely
adopted for improving mathematical reasoning, code generation,
and agentic capabilities in large language models (LLMs)
\citep{shao2024deepseekmath,deepseekr1,lambert2024tulu}. Group-relative
methods obtain their learning signal from multiple trajectories sampled
for the same prompt, making rollout generation a dominant training cost.
The cost grows further for long reasoning traces and agentic tasks.
Asynchronous RL systems improve rollout efficiency by decoupling
generation from policy optimization, with partial rollouts allowing
unfinished trajectories to continue across policy updates
\citep{fu2025areal}. These systems
improve rollout throughput without changing the complete-trajectory
sampling unit of group-relative RL.

Yet tokens within a trajectory need not contribute equally to learning.
\citet{wang2025eighty} show that restricting policy-gradient updates to
the 20\% highest-entropy tokens can match or exceed full-token updates
in their mathematical reasoning experiments. These results suggest that
the learning benefit of a trajectory may depend disproportionately on
a small subset of decisions. However, selecting tokens for optimization
does not reduce the cost of generating the complete trajectories in
the first place. This raises a question at the generation stage: can
the rollout budget be allocated to alternative continuations from
selected positions, while reusing the prefixes that precede them?

\begin{figure}[t]
\centering
\includegraphics[width=\linewidth]{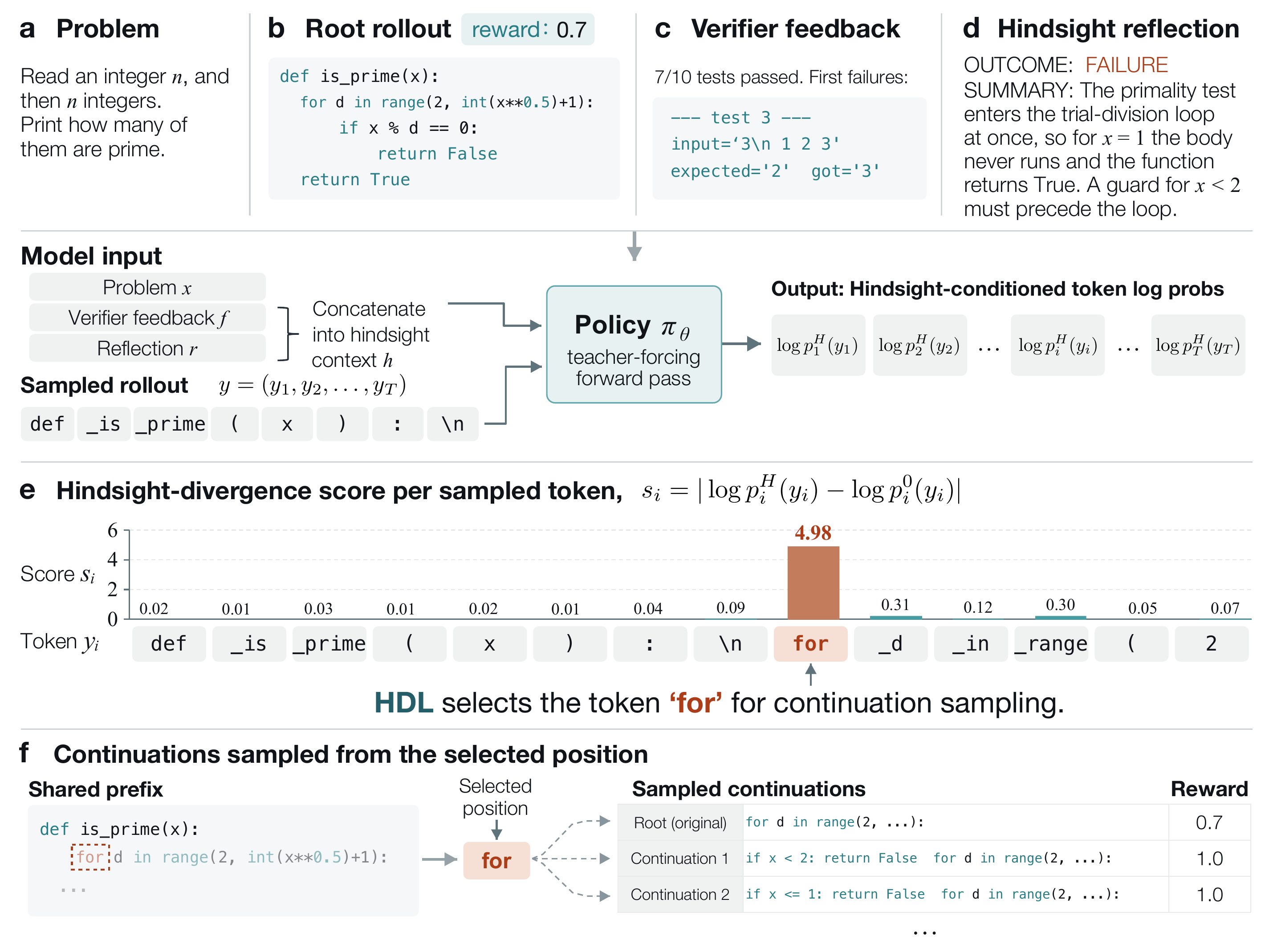}
\caption{\textbf{Hindsight-divergence localization on a coding example.}
The root's primality test incorrectly accepts $1$ as prime. Verifier feedback
prompts a reflection identifying the missing guard for $x < 2$. Hindsight
re-scoring produces the largest absolute log-likelihood change at \texttt{for},
which HDL selects as a branch point. Fresh continuations are sampled under
the original task context while reusing the preceding prefix. The illustrated
continuations introduce alternative guards before the loop, increasing the
reward from 0.7 to 1.0.}
\label{fig:example}
\end{figure}

TreeRL \citep{hou2025treerl} and BPO \citep{bpo2026} use policy uncertainty
to allocate additional rollouts to intermediate decisions, reusing the
preceding prefixes. However, these entropy-based criteria do not use the
observed outcome to reassess earlier decisions. Reflection-based methods
such as PivoARL \citep{pivoarl2026} and R\textsuperscript{3}L \citep{r3l2026}
use completed trajectories and feedback to generate reflections that
explicitly identify where to retry. However, both methods introduce
additional training objectives to develop the model's reflection capability
for retry-point identification.
We evaluate entropy-based and reflection-based methods as baselines in
Section~\ref{sec:localization-signals}.

Recent hindsight self-distillation methods use completed trajectories and their outcomes
to derive token-level supervision, improving performance on reasoning
and agentic tasks \citep{sdsearch2026,hintsd2026,hsd2026}. These methods
compare the model's token predictions with and without hindsight to guide
updates on its own sampled trajectories. The same comparison can also
reveal which earlier choices the model reconsiders after feedback, even
when it was initially confident. This motivates selecting branch points
according to how much hindsight changes the model's assessment of those choices.

We introduce \textbf{Hindsight-Divergence Localization (HDL)}. Given a
completed root trajectory and verifier feedback, the rollout policy
generates a hindsight reflection. HDL re-scores the sampled tokens with
and without a hindsight context containing the feedback and reflection.
The absolute change in each sampled token's log-likelihood defines its
hindsight-divergence score. HDL selects the highest-scoring positions as branch points,
localizing where the model changes its mind after feedback.
Figure~\ref{fig:example} illustrates HDL on a code-generation task:
counting the primes in a list of integers. The root's primality test
incorrectly accepts $1$ as prime. Verifier feedback prompts a reflection
identifying the missing guard for $x < 2$, and the largest log-likelihood
change occurs at \texttt{for}, where the root proceeds to the loop without
this check.

To form a training group, HDL generates a small number of complete roots
and fills the remaining slots with continuations from the selected
positions. Each continuation reuses the corresponding root prefix and
samples a fresh suffix under the original rollout context; hindsight
information is used only for branch selection. In Figure~\ref{fig:example}'s example,
continuations from the selected \texttt{for} position retain the function
definition and explore alternative guards before the loop.
Their verified outcomes provide feedback on alternative choices from
the same history.
Roots and continuations use the same group-relative objective, with
continuation losses restricted to newly generated suffixes. Prefix reuse
reduces generation cost, while the new suffixes concentrate additional
exploration and learning around the selected decisions.

We evaluate HDL with three models across math, code, and agent tasks.
At matched group sizes and training steps, HDL effectively halves the
rollout generation budget relative to GRPO, cutting token usage by up
to 61\% and accelerating wall-clock time by up to 45\%. Task
performance improves across all three domains, with the largest gains
reaching 12.5 points on agent tasks.

\section{Related work}
\label{sec:related}

\paragraph{Efficient RLVR.}
Existing work improves RLVR efficiency through faster rollouts
and selective use of generated data. AReaL decouples generation from
training and supports interruptible rollouts across policy updates
\citep{fu2025areal}; Kimi k1.5 carries unfinished rollouts across training
iterations \citep{kimik15}. DAPO filters groups with uniform rewards and
samples additional groups until the training batch is filled
\citep{yu2025dapo}. Token-selective RL updates only high-entropy tokens
after generating complete trajectories \citep{wang2025eighty}.
HDL reduces generation through prefix reuse while preserving the group
size and RL objective.

\paragraph{Branch-point selection.}
Branching methods reuse prefixes to sample from intermediate states. TreeRL uses
policy uncertainty to guide tree expansion \citep{hou2025treerl}, while
BPO selects high-entropy action states and computes advantages from
sibling returns \citep{bpo2026}. InfoTree combines value estimates,
exploration bonuses, and token entropy to allocate tree expansions
\citep{infotree2026}. PivotRL samples local actions from intermediate
states in existing SFT trajectories and retains turns with mixed outcomes
\citep{pivotrl2026}. PivoARL and R\textsuperscript{3}L use reflection to
identify retry points and guide the regenerated continuations
\citep{pivoarl2026,r3l2026}. HDL derives branch points from changes in
sampled-token log-likelihood rather than explicit error locations.
It samples continuations under the original rollout context without
reflection guidance and retains the group-relative objective.

\paragraph{On-policy self-distillation.}
On-policy self-distillation uses a model conditioned on additional
information to supervise its own sampled trajectories. SDPO conditions
the self-teacher on environment feedback or successful rollouts to obtain
token-level supervision \citep{sdpo2026}. RLSD uses answer-conditioned
token likelihoods to reweight group-relative advantages \citep{rlsd2026},
while SRPO routes trajectories between GRPO and self-distillation
according to their outcomes and the availability of successful peers
\citep{srpo2026}. Other methods tailor hindsight supervision to particular
decisions. SD-Search conditions on
group search traces and outcomes to supervise search-query tokens
\citep{sdsearch2026}. HINT-SD uses full-trajectory hindsight to identify
action spans for feedback-conditioned distillation \citep{hintsd2026}.
HSD uses successful peer trajectories to concentrate supervision near
the divergence from a failed path \citep{hsd2026}. HDL similarly compares
token likelihoods with and without hindsight, but uses their absolute
log-likelihood difference to rank branch points rather than for
token-level supervision.

\section{Hindsight-Divergence Localization}
\label{sec:method}

Group Relative Policy Optimization (GRPO) \citep{shao2024deepseekmath}
samples a group of $G$ complete trajectories $\{y^{(j)}\}_{j=1}^{G}$
independently from the rollout policy $\pi_{\theta_{\mathrm{old}}}$
for each problem $x$. A task verifier assigns each
trajectory a reward $R_j=R(x,y^{(j)})$. GRPO uses these rewards to assess
each trajectory relative to the group. The mean-centered advantage
$A_j=R_j-\frac{1}{G}\sum_{k=1}^{G}R_k$ is positive for trajectories whose
rewards exceed the group mean and negative for those below it.

HDL first samples $M<G$ complete trajectories as roots.
It uses verifier feedback and hindsight reflections to select branch
points within these roots, then fills the remaining $G-M$ slots with
continuations from those positions.

\subsection{Hindsight-conditioned scoring}
\label{sec:hindsight-scoring}

Let $y=(y_1,\ldots,y_T)$ denote one root trajectory and $f$ its verifier
feedback. Given the problem $x$, the completed root $y$, and $f$, the
rollout policy generates a reflection $r$ that interprets the outcome in
relation to earlier decisions. The feedback and reflection form the
hindsight context $h=[f;r]$.

HDL re-scores the root by feeding its recorded tokens back as inputs for
predicting subsequent tokens. This teacher-forced evaluation conditions
the prediction at position $i$ on the original root prefix $y_{<i}$.
For each policy-generated token $y_i$, the next-token distributions under the
original and hindsight-conditioned contexts are
\begin{equation}
\label{eq:hindsight-contexts}
p_i^0(\cdot)
=
\pi_{\theta_{\mathrm{old}}}(\cdot\mid x,y_{<i}),
\qquad
p_i^H(\cdot)
=
\pi_{\theta_{\mathrm{old}}}(\cdot\mid x,h,y_{<i}).
\end{equation}
Both distributions use the same policy parameters and the same root
prefix $y_{<i}$; only the hindsight context differs.

\subsection{Branch-point selection}
\label{sec:localization}

To identify decisions whose assessment changes under hindsight, HDL
scores each sampled token $y_i$ by the absolute change in its
log-likelihood:
\begin{equation}
\label{eq:div}
s_i
=
\left|
\log p_i^H(y_i)-\log p_i^0(y_i)
\right|.
\end{equation}
We refer to $s_i$ as the \emph{hindsight-divergence score}.
After the outcome is known, hindsight may increase the
likelihood of tokens at key steps in a successful trajectory. In a failed
trajectory, it may decrease the likelihood of tokens at a step where an
error occurred. Taking the absolute value captures both increased and
decreased support for the sampled token.

HDL ranks candidate positions within each root by $s_i$
and selects the highest-scoring positions as branch points.

\subsection{Localized group construction}
\label{sec:fork}

The selected branch points determine where to sample the remaining
$G-M$ trajectories. Given a root $y$ and branch point $i$, HDL reuses
the prefix $y_{<i}$ and samples a fresh suffix under the original task
context:
\begin{equation}
\label{eq:branch-sampling}
\widetilde y_{\ge i}
\sim
\pi_{\theta_{\mathrm{old}}}(\cdot\mid x,y_{<i}).
\end{equation}
The prefix and suffix form a complete trajectory
$\widetilde y=y_{<i}\Vert\widetilde y_{\ge i}$.
Section~\ref{sec:setup} specifies the default root count $M$ and allocation
of the $G-M$ continuations across roots and branch points;
Section~\ref{sec:branching-configs} compares alternative branching configurations.

The $M$ roots and $G-M$ continuations form a single training group,
whose verifier rewards determine the advantages $A_j$ defined above.
The policy is optimized with the same objective as the GRPO baseline.
Each root contributes policy loss over its generated tokens. For a
continuation, the reused prefix provides context, and the loss is applied
only to newly sampled tokens. This avoids counting the shared prefix
again in each continuation's loss and focuses its learning signal on
the decisions explored after branching.

\section{Evaluation}
\label{sec:results}

\subsection{Experimental setup}
\label{sec:setup}

\paragraph{Tasks and training data.} We study three domains with
verifiable outcomes: Math, Code, and Agent.
\begin{itemize}
\item \textbf{Math.} We draw problems from DeepMath-103K
\citep{he2025deepmath}. Before training, we use the initial policy to
filter out problems that are either too easy or too difficult,
retaining 4{,}555 unique problems. Exact-answer verification provides a
binary reward. Feedback consists of a correctness verdict and, for
incorrect solutions, the predicted and reference answers.
\item \textbf{Code.} We combine the TACO and PrimeIntellect subsets of
DeepCoder \citep{deepcoder2025} with the \texttt{seed\_testcase} subset of
rStar-Coder \citep{liu2025rstarcoder}. Applying the same filtering
procedure leaves 4{,}063 unique problems. Programs are executed against
stdin/stdout tests; the reward is the fraction of tests passed, and the
feedback reports the pass count and details of the first failing tests.
\item \textbf{Agent.} We use ScienceWorld \citep{wang2022scienceworld},
a text-based interactive environment in which an agent completes
elementary-science tasks by navigating rooms and manipulating objects
through natural-language actions. Its training split contains 1{,}856
task--variation pairs across 30 task types after limiting each type to at
most 200 variations. Episodes are limited to 30 actions. The reward is
the environment's cumulative subgoal score normalized to $[0,1]$, and
the feedback reports the final score and whether the task was completed.
\end{itemize}

\paragraph{Models.} We evaluate HDL on Qwen3-4B, Qwen3-8B
\citep{qwen3}, and Llama-3.1-Nemotron-Nano-8B-v1
\citep{nemotronnano}, abbreviated as Llama3.1-8B.
Qwen3 uses thinking mode for Math and Code and non-thinking mode for
Agent; Llama3.1-8B uses its reasoning system prompt. Hindsight reflections
are generated in non-thinking mode for all models
(prompt template in Appendix~\ref{sec:hdl-reflection-prompt}).

\paragraph{Training.} All experiments use the slime framework
\citep{slime_github} on four nodes, each equipped with four GB200 GPUs.
All methods share the same training settings: 128 problems per step,
$G{=}16$ trajectories per group, 200 optimization steps, learning rate
$10^{-6}$, and sampling temperature 1.0. Math and Code responses are
limited to 32{,}768 tokens. Agent trajectories, including environment
observations, are limited to 8{,}192 tokens for Qwen3 and 16{,}384 for
Llama3.1. The same length limits apply during evaluation.
We train all methods with GRPO, omitting group standard-deviation
normalization following Dr.\ GRPO \citep{liu2025drgrpo}. Losses are aggregated
at the token level. We use asymmetric clipping with lower and upper
thresholds of 0.2 and 0.28, respectively, and no KL penalty.

\paragraph{Rollout protocols.}
\label{sec:selectors}
GRPO independently samples $G{=}16$ complete trajectories per problem.
DAPO uses dynamic sampling \citep{yu2025dapo}, filtering out groups
with uniform rewards and sampling additional complete trajectories
to replace them.

HDL samples $M{=}2$ complete roots per problem. For each root, it selects
the two positions with the highest hindsight-divergence scores and
allocates three and four continuations to these positions. Including
the roots, this gives $G{=}2\times(1+3+4){=}16$ trajectories per group.
Each continuation reuses its root prefix and samples a fresh suffix
under the original rollout context. We compare alternative branching
configurations in Section~\ref{sec:branching-configs}.

For the localization-signal comparison in
Section~\ref{sec:localization-signals}, we compare HDL with two
alternative localization methods:

\emph{Entropy} follows the use of policy entropy to guide branching in
TreeRL \citep{hou2025treerl} and BPO \citep{bpo2026}. It ranks candidate
positions by the entropy of the next-token distribution conditioned
on the problem and root prefix, without verifier feedback or hindsight
context.

\emph{Reflection} uses explicit self-reflection to identify retry points,
as in PivoARL \citep{pivoarl2026} and R\textsuperscript{3}L
\citep{r3l2026}. Given the completed root and verifier feedback, the
policy is prompted to identify the earliest erroneous step, or a step
worth revisiting when the root is successful
(prompt templates in Appendix~\ref{sec:reflection-baseline-prompt}).
The returned step indices are mapped to branch points.

Both alternatives use the same root--continuation allocation,
original-context continuation sampling, and RL objective as HDL.

\paragraph{Evaluation.} For task performance, we evaluate mathematical reasoning on AIME24
\citep{mathai2024aime}, AIME25 \citep{mathai2025aime}, AIME26
\citep{mathai2026aime}, HMMT February 2026 \citep{matharena2026hmmt},
Minerva Math \citep{lewkowycz2022minerva}, and
OlympiadBench \citep{he2024olympiadbench}; code generation on LiveCodeBench
v5 and v6 \citep{jain2024livecodebench}; and agent performance on held-out
ScienceWorld task variations \citep{wang2022scienceworld}. For Math and
Code, we report average accuracy over $N$ independently sampled responses
per problem (avg@$N$): $N{=}16$ for AIME and HMMT, $N{=}8$ for Minerva Math
and LiveCodeBench, and $N{=}4$ for OlympiadBench. For Agent, we average
task scores over four episodes per held-out variation. Evaluation is performed every 10 training steps with
a sampling temperature of 0.6.
For each method and model, we select the three checkpoints with the
highest average benchmark score within each domain and report the mean
and standard deviation of each metric across these checkpoints.

For rollout efficiency, we report generated tokens and end-to-end rollout
wall-clock time per training step. Token counts include HDL's hindsight
reflections and candidates discarded by DAPO's dynamic sampling.
Wall-clock time covers the full rollout pipeline, including reflection
generation and hindsight scoring for HDL.

\subsection{Rollout efficiency}
\label{sec:efficiency}

\begin{figure*}[t]
\centering
\includegraphics[width=\textwidth]{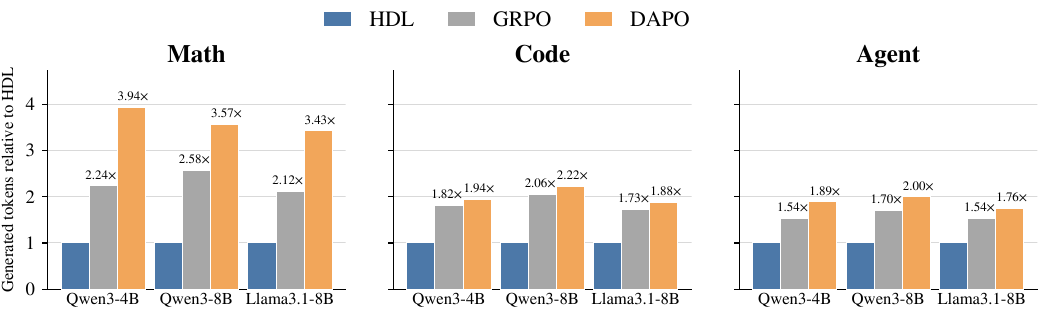}
\caption{\textbf{HDL reduces generated tokens by 35--61\% relative to
GRPO.} Bars show mean generated tokens per training step, normalized to
HDL within each model--task pair; annotations give the corresponding
cost ratios.}
\label{fig:tokens}
\end{figure*}

\begin{figure*}[t]
\centering
\includegraphics[width=\textwidth]{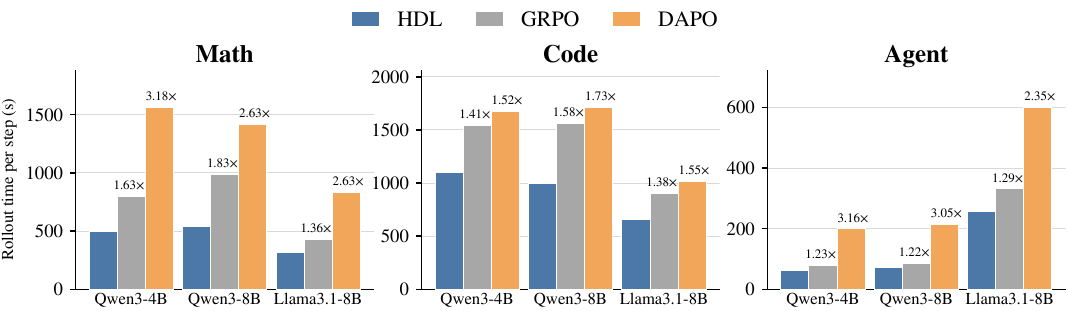}
\caption{\textbf{HDL reduces end-to-end rollout time by 18--45\% relative
to GRPO.} Bars show mean seconds per training step on identical hardware;
annotations give time ratios relative to HDL. Each task uses a separate
vertical scale.}
\label{fig:walltime}
\vspace{-4pt}
\end{figure*}

HDL cuts generated tokens by 35--61\% and end-to-end rollout wall-clock
time by 18--45\% relative to GRPO. These savings hold across all three
models on Math, Code,
and Agent tasks. Figures~\ref{fig:tokens} and \ref{fig:walltime} report
the corresponding costs per training step.

\paragraph{Generated tokens.}
The largest reductions relative to GRPO occur on Math, where HDL more
than halves generation for every model (Figure~\ref{fig:tokens}).
For Qwen3-8B, mean generation falls from 27.28M to 10.59M tokens per
training step. Compared with DAPO, HDL reduces generated tokens by
43--75\% across the three domains.

\paragraph{Rollout wall-clock time.}
The generation savings translate into faster rollout collection for
every model and task (Figure~\ref{fig:walltime}). HDL achieves
$1.22\times$--$1.83\times$ rollout speedups over GRPO and
$1.52\times$--$3.18\times$ over DAPO, corresponding to 34--69\% less
rollout time than DAPO.
Thus, prefix reuse yields substantial savings in the complete rollout
pipeline even after accounting for HDL's localization overhead.

\subsection{Downstream task performance}
\label{sec:eval-results}

HDL delivers task-performance gains alongside its rollout savings,
reaching up to 12.46 percentage points over GRPO on Agent tasks
(Tables~\ref{tab:math}--\ref{tab:agent}).

\newcommand{\evalscore}[2]{\ensuremath{#1_{\scriptscriptstyle\pm #2}}}

\begin{table}[t]
\centering
\caption{\textbf{Math.} Scores (\%, $\uparrow$) on six benchmarks and
their unweighted average (Avg.). Entries show mean$_{\pm\mathrm{std}}$
over the three checkpoints with the highest Avg. Bold and underlining
mark the best and second-best means within each model, respectively.}
\label{tab:math}
\small
\setlength{\tabcolsep}{3.5pt}
\begin{tabular}{lccccccc}
\toprule
\multicolumn{1}{c}{Method} & \multicolumn{1}{c}{AIME24} &
\multicolumn{1}{c}{AIME25} & \multicolumn{1}{c}{AIME26} &
\multicolumn{1}{c}{HMMT} & \multicolumn{1}{c}{Minerva} &
\multicolumn{1}{c}{Olympiad} & \multicolumn{1}{c}{Avg.} \\
\midrule
\multicolumn{8}{@{}l}{\textit{Qwen3-4B}} \\
GRPO & \evalscore{\mathbf{74.51}}{0.26} & \evalscore{65.62}{0.74} & \evalscore{66.46}{0.29} & \evalscore{\underline{19.44}}{0.39} & \evalscore{\underline{31.13}}{0.45} & \evalscore{52.25}{0.16} & \evalscore{51.57}{0.07} \\
DAPO & \evalscore{\underline{74.51}}{0.60} & \evalscore{\mathbf{68.82}}{0.35} & \evalscore{\mathbf{68.96}}{0.51} & \evalscore{18.37}{0.54} & \evalscore{\mathbf{31.20}}{0.32} & \evalscore{\underline{53.04}}{0.10} & \evalscore{\mathbf{52.48}}{0.02} \\
\rowcolor{black!7}
HDL & \evalscore{73.12}{0.68} & \evalscore{\underline{67.01}}{0.80} & \evalscore{\underline{67.57}}{1.52} & \evalscore{\mathbf{19.89}}{1.07} & \evalscore{30.38}{0.21} & \evalscore{\mathbf{53.18}}{0.45} & \evalscore{\underline{51.86}}{0.14} \\
\midrule
\multicolumn{8}{@{}l}{\textit{Qwen3-8B}} \\
GRPO & \evalscore{\underline{75.62}}{0.34} & \evalscore{68.40}{0.64} & \evalscore{68.33}{2.78} & \evalscore{21.09}{1.17} & \evalscore{\mathbf{32.72}}{0.10} & \evalscore{52.86}{0.29} & \evalscore{53.17}{0.13} \\
DAPO & \evalscore{74.86}{0.69} & \evalscore{\underline{69.58}}{0.34} & \evalscore{\mathbf{69.17}}{0.74} & \evalscore{\underline{21.46}}{1.20} & \evalscore{\underline{32.57}}{0.18} & \evalscore{\underline{53.20}}{0.08} & \evalscore{\underline{53.47}}{0.13} \\
\rowcolor{black!7}
HDL & \evalscore{\mathbf{75.97}}{0.87} & \evalscore{\mathbf{70.56}}{0.35} & \evalscore{\underline{69.10}}{0.55} & \evalscore{\mathbf{23.30}}{0.15} & \evalscore{32.49}{0.14} & \evalscore{\mathbf{53.52}}{0.35} & \evalscore{\mathbf{54.16}}{0.10} \\
\midrule
\multicolumn{8}{@{}l}{\textit{Llama3.1-8B}} \\
GRPO & \evalscore{\underline{68.68}}{0.94} & \evalscore{\underline{55.76}}{0.98} & \evalscore{\underline{65.69}}{0.55} & \evalscore{\underline{27.90}}{0.50} & \evalscore{\underline{29.20}}{0.18} & \evalscore{61.70}{0.87} & \evalscore{\underline{51.49}}{0.34} \\
DAPO & \evalscore{\mathbf{70.35}}{0.49} & \evalscore{\mathbf{59.86}}{0.64} & \evalscore{\mathbf{68.75}}{0.88} & \evalscore{27.59}{1.35} & \evalscore{\mathbf{29.81}}{0.31} & \evalscore{\mathbf{63.69}}{0.43} & \evalscore{\mathbf{53.34}}{0.45} \\
\rowcolor{black!7}
HDL & \evalscore{68.19}{0.60} & \evalscore{55.28}{1.16} & \evalscore{64.03}{1.13} & \evalscore{\mathbf{29.04}}{1.18} & \evalscore{28.86}{0.25} & \evalscore{\underline{62.95}}{0.55} & \evalscore{51.39}{0.19} \\
\bottomrule
\end{tabular}
\end{table}

\begin{table}[t]
\centering
\begin{minipage}[t]{0.49\linewidth}
\centering
\caption{\textbf{Code.} Avg@8 accuracy on
LCB v5 and v6 (\%, $\uparrow$).}
\label{tab:code}
\small
\setlength{\tabcolsep}{2pt}
\begin{tabular*}{\linewidth}{@{\extracolsep{\fill}}lcc>{\columncolor{black!7}[\tabcolsep][0pt]}c@{}}
\toprule
\multicolumn{1}{c}{Model} & \multicolumn{1}{c}{GRPO} &
\multicolumn{1}{c}{DAPO} &
\multicolumn{1}{>{\columncolor{black!7}}c}{HDL} \\
\midrule
Qwen3-4B & \evalscore{\underline{54.02}}{0.12} & \evalscore{53.55}{0.09} & \evalscore{\mathbf{54.15}}{0.10} \\
Qwen3-8B & \evalscore{\underline{55.42}}{0.22} & \evalscore{54.88}{0.08} & \evalscore{\mathbf{55.56}}{0.09} \\
Llama3.1-8B & \evalscore{\mathbf{54.87}}{0.43} & \evalscore{\underline{54.48}}{0.47} & \evalscore{54.34}{0.29} \\
\bottomrule
\end{tabular*}
\end{minipage}\hfill
\begin{minipage}[t]{0.49\linewidth}
\centering
\caption{\textbf{Agent.} Mean task-completion
score on ScienceWorld (\%, $\uparrow$).}
\label{tab:agent}
\label{tab:sci}
\small
\setlength{\tabcolsep}{2pt}
\begin{tabular*}{\linewidth}{@{\extracolsep{\fill}}lcc>{\columncolor{black!7}[\tabcolsep][0pt]}c@{}}
\toprule
\multicolumn{1}{c}{Model} & \multicolumn{1}{c}{GRPO} &
\multicolumn{1}{c}{DAPO} &
\multicolumn{1}{>{\columncolor{black!7}}c}{HDL} \\
\midrule
Qwen3-4B & \evalscore{57.76}{0.45} & \evalscore{\underline{60.84}}{0.29} & \evalscore{\mathbf{67.44}}{1.42} \\
Qwen3-8B & \evalscore{59.50}{0.15} & \evalscore{\underline{60.80}}{0.52} & \evalscore{\mathbf{71.96}}{1.55} \\
Llama3.1-8B & \evalscore{69.94}{1.22} & \evalscore{\mathbf{75.35}}{0.86} & \evalscore{\underline{71.81}}{0.97} \\
\bottomrule
\end{tabular*}
\end{minipage}
\par\smallskip
\begin{minipage}{\linewidth}
\footnotesize Entries report mean$_{\pm\mathrm{std}}$ over the three
checkpoints with the highest average score in each domain. Bold and
underlining indicate the best and second-best means within each model.
\end{minipage}
\end{table}

\paragraph{Mathematical reasoning and code generation.}
HDL outperforms GRPO on both Qwen3 models in Math and Code.
On Qwen3-8B Math, HDL achieves the highest average score (54.16\%), outperforming
both GRPO (53.17\%) and DAPO (53.47\%).
On Qwen3-4B, HDL scores 51.86\%, outperforming GRPO and remaining within
0.62 points of DAPO, despite DAPO consuming nearly four times as many tokens.
On Code tasks, HDL achieves the highest accuracy on both Qwen3-4B (54.15\%) and Qwen3-8B (55.56\%).
HDL achieves these results with substantially fewer generated tokens.

\paragraph{Agent.}
The most pronounced performance gains emerge in the Agent domain (ScienceWorld),
where multi-step sequential execution creates challenging credit assignment problems.
HDL improves over standard GRPO by 9.68 points on Qwen3-4B (67.44\% vs 57.76\%) and by
12.46 points on Qwen3-8B (71.96\% vs 59.50\%), while also outperforming DAPO by 6.60
and 11.16 points, respectively.

In long-horizon interactive environments, early sub-optimal actions (such as navigating to
an incorrect room or selecting the wrong tool) cascade into irreversible failure, causing
independently sampled rollouts to redundantly explore failed trajectories from scratch.
By localizing the critical turning points in hindsight and branching multiple fresh suffixes,
HDL effectively rescues near-failure episodes.
This creates high-contrast advantage groups with informative reward variance, accelerating
policy improvement on interactive decision-making tasks.

\Needspace{8\baselineskip}
\subsection{Comparison of localization signals}
\label{sec:localization-signals}

To evaluate the choice of localization signal, we compare HDL with Entropy
and Reflection on Qwen3-8B under the same training settings,
root--continuation allocation, and branch-point constraints.
HDL achieves the highest scores in all three domains
(Table~\ref{tab:selectors}). Its advantage is largest on Agent tasks,
where it exceeds Entropy and Reflection by 5.67 and 6.54 percentage
points, respectively. We examine their branch-point selections on
ScienceWorld to understand this difference.

\Needspace{14\baselineskip}
\begin{wrapfigure}{r}{0.41\textwidth}
\vspace{-8pt}
\centering
\input{figs/sci_interaction.tex}
\setlength{\abovecaptionskip}{4pt}
\caption{\raggedright\textbf{A single ScienceWorld interaction step.}}
\label{fig:sci-interaction}
\vspace{-8pt}
\end{wrapfigure}

Figure~\ref{fig:sci-interaction} illustrates a single interaction step
in ScienceWorld. The agent observes the current state, generates an
action, and receives feedback from the environment. Within the action,
the opening verb \texttt{connect} specifies the operation, while its
arguments identify the bulb's cathode and the battery's anode.

\begin{table}[t]
\centering
\vspace{-4pt}
\caption{\textbf{Localization-signal comparison on Qwen3-8B.}
Scores (\%, $\uparrow$) are reported as mean$_{\pm\mathrm{std}}$ over the
three checkpoints with the highest average score in each domain.
Bold and underlining mark the best and second-best means.}
\label{tab:selectors}
\small
\setlength{\tabcolsep}{10pt}
\begin{tabular}{l>{\columncolor{black!7}}ccc}
\toprule
\multicolumn{1}{c}{Task} &
\multicolumn{1}{>{\columncolor{black!7}}c}{HDL} &
Entropy & Reflection \\
\midrule
Math & \evalscore{\mathbf{54.16}}{0.10} & \evalscore{\underline{54.02}}{0.25} & \evalscore{53.12}{0.05} \\
Code & \evalscore{\mathbf{55.56}}{0.09} & \evalscore{\underline{55.37}}{0.26} & \evalscore{53.95}{0.10} \\
Agent & \evalscore{\mathbf{71.96}}{1.55} & \evalscore{\underline{66.29}}{1.59} & \evalscore{65.42}{0.66} \\
\bottomrule
\end{tabular}
\end{table}

\paragraph{Comparison with Entropy.}
\begin{figure}[t]
\centering
\includegraphics[width=1.0\linewidth]{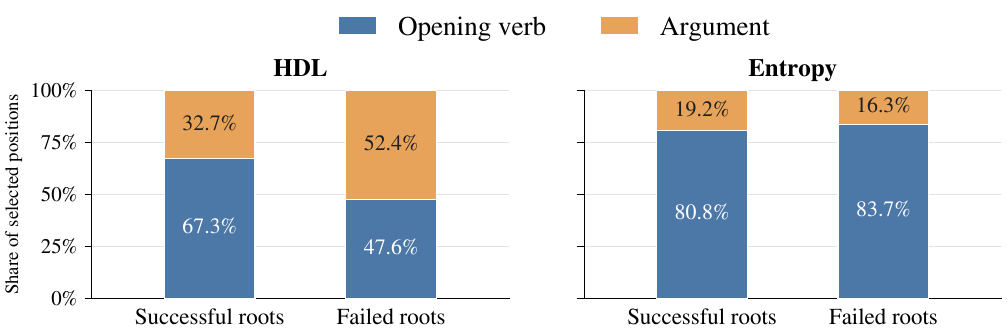}
\caption{\textbf{Proportions of HDL and Entropy branch points on opening verbs
and arguments in successful and failed roots.}}
\label{fig:localization-outcomes}
\end{figure}

Entropy's preference for opening verbs is nearly unchanged by root
outcome (Figure~\ref{fig:localization-outcomes}): 80.8\% of its branch
points fall on verbs in successful roots and 83.7\% in failed roots.
This bias reflects action structure: many operations
compete at the opening verb, while choosing one constrains the arguments
that follow. HDL, by contrast, places more branch points on arguments
in failed roots than in successful roots (52.4\% vs. 32.7\%), allowing
continuations to vary the object or destination while retaining the
operation.

\begin{figure}[t]
\centering
\includegraphics[width=1.0\linewidth]{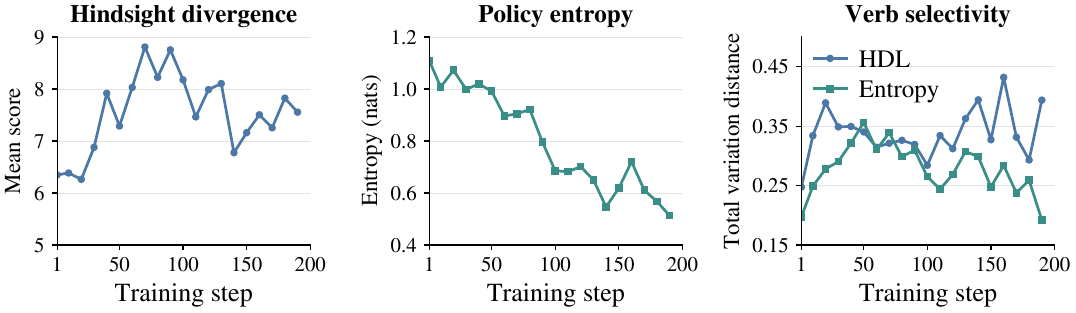}
\caption{\textbf{Localization signals and verb selectivity.}
Left and middle: mean signal at positions selected by HDL and Entropy, respectively.
Right: verb selectivity, measured by total variation distance (TVD).
Higher values indicate stronger preferences for particular opening verbs.}
\label{fig:localization-training}
\end{figure}

Entropy also weakens during training (Figure~\ref{fig:localization-training}).
Mean entropy at its selected positions falls from 1.03 in steps 1--50 to
0.57 in steps 151--200. Over the same windows, Entropy's selected verbs
move closer to the distribution of all opening verbs in the same roots
(TVD 0.278 to 0.246), so its selections increasingly track verb
frequency rather than a distinct subset. HDL's mean hindsight-divergence
score rises from 6.85 to 7.78, while its selected verbs remain more
distinct from the root distribution (TVD 0.329 to 0.348).

\paragraph{Comparison with Reflection.}
Reflection prompts the model to specify revisit points directly.
Some returned position identifiers cannot be parsed or mapped to valid
positions. More often, the proposed points are too close together,
so only one is retained. Reflection consequently yields an
average of 3.22 branch points per group, compared with 3.95 for HDL,
out of a maximum of four. Even when only one valid branch point remains
for a root, we sample all of its allocated continuations from that
point to maintain the same group size.

\subsection{Comparison of branching configurations}
\label{sec:branching-configs}

With HDL as the localization signal, we compare three branching
configurations on Qwen3-8B Agent at the same group size
(Table~\ref{tab:fork-ablation}). Each configuration is written as
roots $\times$ branch points per root, with $2\times2$ as the default.

\begin{table}[t]
\centering
\vspace{-4pt}
\caption{\textbf{Branching configurations on Qwen3-8B Agent.}
Points are counted per root. Tokens and rollout time are per-step means.}
\label{tab:fork-ablation}
\small
\setlength{\tabcolsep}{7pt}
\begin{tabular}{lcccc}
\toprule
\multicolumn{1}{c}{Roots $\times$ Points} &
\shortstack{Continuations\\per root} &
Score (\%) $\uparrow$ & Tokens (M) $\downarrow$ & Time (s) $\downarrow$ \\
\midrule
\rowcolor{black!7}
$2\times2$ (default) & $3+4$ & $\mathbf{71.96}_{\pm1.55}$ & 0.27 & 70.46 \\
$2\times1$ & $7$ & $71.09_{\pm0.29}$ & \textbf{0.25} & \textbf{59.97} \\
$4\times2$ & $1+2$ & $67.87_{\pm0.44}$ & 0.31 & 68.07 \\
\bottomrule
\end{tabular}
\end{table}

\vspace{-8pt}
The default $2\times2$ achieves the highest score (71.96\%). Reducing
the number of branch points per root to one ($2\times1$) concentrates
all seven continuations at a single position. This reduces rollout
time by 15\%, at a 0.87-point decrease in task score.

Increasing the number of roots to four ($4\times2$) leaves only one
or two continuations per branch point. This configuration generates
more tokens and scores 4.09 points below the default. Together, these
comparisons favor revisiting multiple positions within each root while
retaining several continuations per position.

\vspace{-0.5em}
\section{Conclusion}
\label{sec:conclusion}
We introduced Hindsight-Divergence Localization (HDL) to allocate rollout
generation to intermediate decisions that the model reconsiders after
feedback. HDL identifies these positions through hindsight-induced
changes in token log-likelihoods, then builds training groups from a
small number of complete roots and continuations sampled under the
original task context. Prefix reuse reduces generation cost, while the
new suffixes concentrate additional exploration and learning around the
selected decisions. Experiments with three models across math, code,
and agent tasks show that HDL reduces generated tokens by 35--61\% and
rollout wall-clock time by 18--45\% relative to GRPO, while improving
task performance across all three domains, with gains of up to 12.5
percentage points on agent tasks. In controlled comparisons, HDL also
achieves higher task scores than entropy-based and reflection-based
localization methods, supporting hindsight divergence as a criterion
for deciding where to branch.

\bibliography{divrl}
\bibliographystyle{references}

\clearpage
\appendix
\section{Limitations}
\label{sec:limitations}

HDL relies on the premise that the policy can reliably interpret verifier feedback to
re-score its own decisions.
We examine the boundary of this capability across model scales.
In its reflection, the policy reports whether the root trajectory succeeded or failed.
Table~\ref{tab:label-agreement} reports the percentage of reflections in which this
outcome matches the verifier's verdict.
On Qwen3-8B, the policy achieves near-perfect outcome reporting (99.7--99.9\%).
However, on Qwen3-1.7B, outcome agreement drops sharply to 76.0\% on Math and 56.1\% on Code.
Even when Math feedback consists of a single word (``Correct.''), the 1.7B policy frequently
hallucinates or misattributes the verdict.

\begin{table}[htbp]
\centering
\begin{minipage}[t]{0.44\linewidth}
\vspace{0pt}
\centering
\captionof{table}{\textbf{Outcome-label agreement.} Percentage of reflections whose
success/failure label matches the verifier verdict (\%, $\uparrow$), on identical training pools.}
\label{tab:label-agreement}
\small
\setlength{\tabcolsep}{7pt}
\begin{tabular}{lcc}
\toprule
Model & Math & Code \\
\midrule
Qwen3-8B & 99.7 & 99.9 \\
Qwen3-1.7B & \textbf{76.0} & \textbf{56.1} \\
\bottomrule
\end{tabular}
\end{minipage}\hfill
\begin{minipage}[t]{0.54\linewidth}
\vspace{0pt}
\centering
\captionof{table}{\textbf{ScienceWorld across model scales.} Scores (\%, $\uparrow$) are
mean$_{\pm\mathrm{std}}$ over the top three checkpoints. Bold marks the
best-performing method for each model.}
\label{tab:scale-sci}
\small
\setlength{\tabcolsep}{2.5pt}
\begin{tabular}{lccc}
\toprule
Model & GRPO & Entropy & HDL \\
\midrule
Qwen3-1.7B & \evalscore{45.66}{0.75} & \evalscore{\mathbf{47.54}}{0.85} & \evalscore{45.70}{0.19} \\
Qwen3-8B & \evalscore{59.50}{0.15} & \evalscore{66.29}{1.59} & \evalscore{\mathbf{71.96}}{1.55} \\
\bottomrule
\end{tabular}
\end{minipage}
\end{table}

As shown in Table~\ref{tab:scale-sci}, on Qwen3-1.7B ScienceWorld, HDL's performance advantage
vanishes, matching GRPO (45.70\% vs 45.66\%) and trailing Entropy (47.54\%).
When the model cannot reliably understand its own feedback, hindsight re-scoring can
introduce noise into branch-point ranking.

\section{Reflection prompts}
\label{sec:reflection-prompts}

\subsection{HDL reflection}
\label{sec:hdl-reflection-prompt}

HDL uses the following prompt across Math, Code, and Agent tasks.
Braced fields contain the problem, root trajectory, and verifier feedback.

\begin{quote}
\small\ttfamily\raggedright
\{problem\}\par\medskip
[A completed attempt]\\
\{root\}\par\medskip
[Record]\\
\{verifier feedback\}\par\medskip
Summarize this attempt in your own words: what approach it took, and why,
according to the record, it turned out the way it did. If it went wrong,
say what the right approach would have been. Do not quote the attempt
verbatim. Do not mention positions, line numbers or percentages.\par\medskip
Answer in exactly this format (60-120 tokens for the summary):\par\medskip
OUTCOME: SUCCESS or FAILURE\\
SUMMARY: <your summary>\par\medskip
Stop immediately after the summary: write nothing after it.
\end{quote}

\subsection{Reflection baseline}
\label{sec:reflection-baseline-prompt}

The Reflection baseline requests two branch points directly. Math and
Code roots are presented as numbered steps; Agent roots are presented
as numbered interaction turns. The fields \texttt{lo} and \texttt{hi}
specify the allowed index range.

\paragraph{Math and Code.}
\begin{quote}
\small\ttfamily\raggedright
You attempted this problem:\\
\{problem\}\par\medskip
Your attempt, split into numbered steps:\\
\{numbered steps\}\par\medskip
Feedback: \{verifier feedback\}\par\medskip
\{selection instruction\} Then give a second, different step as an
alternative. Choose steps between \{lo\} and \{hi\}. Answer with exactly
two lines:\\
STEP: <number>\\
STEP2: <number>
\end{quote}

For failed roots, the selection instruction is:
\begin{quote}
\small\ttfamily\raggedright
Identify the EARLIEST step where the attempt goes wrong.
\end{quote}
For successful roots, it is:
\begin{quote}
\small\ttfamily\raggedright
Identify the step where you would BRANCH to explore a different,
potentially better continuation.
\end{quote}

\paragraph{Agent.}
\begin{quote}
\small\ttfamily\raggedright
You attempted this task:\\
\{problem\}\par\medskip
Your episode, split into numbered turns:\\
\{numbered turns\}\par\medskip
Feedback: \{verifier feedback\}\par\medskip
\{selection instruction\} Then give a second, different turn as an
alternative. Choose turns between \{lo\} and \{hi\}. Answer with exactly
two lines:\\
STEP: <number>\\
STEP2: <number>
\end{quote}

For failed roots, the selection instruction is:
\begin{quote}
\small\ttfamily\raggedright
Identify the EARLIEST turn where the episode goes wrong.
\end{quote}
For successful roots, it is:
\begin{quote}
\small\ttfamily\raggedright
Identify the turn where you would BRANCH to try a different,
potentially better course of action.
\end{quote}

\end{document}